\pdfoutput=1

\documentclass[11pt]{article}

\usepackage[preprint]{acl}
\usepackage{multirow} 
\usepackage{array} 
\usepackage{multicol}

\usepackage{times}
\usepackage{latexsym}
\usepackage{makecell}
\usepackage{subfigure} 

\usepackage[T1]{fontenc}

\usepackage[utf8]{inputenc}

\usepackage{microtype}

\usepackage{inconsolata}

\usepackage{graphicx}
\usepackage{booktabs} 
\usepackage[most]{tcolorbox}
\tcbuselibrary{skins, breakable, theorems}
\tcbuselibrary{listings, breakable}
\usepackage{adjustbox}
\usepackage{tablefootnote}
\usepackage{cuted}
\usepackage{lipsum}

\title{TableReasoner: Advancing Table Reasoning Framework with Large Language Models}

\author{
 \textbf{Sishi Xiong\textsuperscript{}}, 
 \textbf{Dakai Wang\textsuperscript{}}, 
 \textbf{Yu Zhao\textsuperscript{}}, 
 \textbf{Jie Zhang\textsuperscript{}}, 
\\
 \textbf{Changzai Pan\textsuperscript{}},
 \textbf{Haowei He\textsuperscript{}},
 \textbf{Xiangyu Li\textsuperscript{}},
 \textbf{Wenhan Chang\textsuperscript{}},
 \\
 \textbf{Zhongjiang He\textsuperscript{}},
 \textbf{Shuangyong Song$^*$\textsuperscript{}},
 \textbf{Yongxiang Li$^*$\textsuperscript{}}
\\
\\
 \textsuperscript{}Institute of Artificial Intelligence (TeleAI), China Telecom Corp Ltd
\\
 \small{
   \textbf{Correspondence:} \href{mailto:email@domain}{\{xiongsishi, songshy, liyx25\}@chinatelecom.cn}
 }
}

\begin{document}
\maketitle
\begin{abstract}
The paper presents our system developed for table question answering (TQA). TQA tasks face challenges due to the characteristics of real-world tabular data, such as large size, incomplete column semantics, and entity ambiguity. To address these issues, we propose a large language model (LLM)-powered and programming-based table reasoning framework, named TableReasoner. It models a table using the schema that combines structural and semantic representations, enabling holistic understanding and efficient processing of large tables. We design a multi-step schema linking plan to derive a focused table schema that retains only query-relevant information, eliminating ambiguity and alleviating hallucinations. This focused table schema provides precise and sufficient table details for query refinement and programming. Furthermore, we integrate the reasoning workflow into an iterative thinking architecture, allowing incremental cycles of thinking, reasoning and reflection. Our system achieves first place in both subtasks of SemEval-2025 Task 8 \footnote{Codes: \url{https://github.com/ccx06/TableReasoner}.}.

\end{abstract}

\section{Introduction}

\newcommand\blfootnote[1]{%
\begingroup
\renewcommand\thefootnote{}\footnote{#1}%
\addtocounter{footnote}{-1}%
\endgroup
}
\blfootnote{* Corresponding authors.}


Table Question Answering is a spin-off Question Answering (QA) task, aiming at responding to natural language questions based on table data that feature heterogeneous and complex two-dimensional structure \citep{lu2025large}.
Compared to tasks involving unstructured or plain text, TQA faces greater challenges and emphasizes the model's reasoning abilities. The primary challenges encompass large size, incomplete semantics, and entity ambiguity. To advance research on table understanding by applying language models, SemEval-2025 introduces Task 8: Question Answering on Tabular Data \citep{osesgrijalba-etal-2025-semeval-2025}.


In this paper, we present TableReasoner, a systematic LLM-powered and programming-based framework for TQA. We introduce the table schema as the table representation, instead of entire or truncated table text in conventional format (such as CSV or Markdown), enabling our framework capable of processing large tables. TableReasoner employs a programming module to solve questions using the table schema to understand the table content from a holistic perspective. To ensure precise scheduling for query decomposition and programming, we design a "parsing-linking-refinement" action flow. It refines the global table schema into a focused one only associated with the original query, to some extent alleviating model hallucinations. 
In addition, inspired by the ReAct \citep{yao2023react} paradigm, we incorporate the reasoning workflow into a "thought-action-observation" architecture, facilitating iterative cycles of thinking, reasoning and reflection within our system.

TableReasoner is flexible to any advanced language model. It delivers excellent results even without fine-tuning, while further performance enhancements can be achieved through fine-tuning and majority voting. Our system wins first place on both subtasks, validating the superiority and scalability of our framework. 

\section{Background}
\newcommand{\firstsentencebold}[1]{%
    \noindent%
    \textbf{\ignorespaces#1}
    \unskip\unskip
    \ignorespacesafterend
}


\subsection{DataBench Dataset}
DataBench \citep{oses-etal-2024-databench} is an English benchmark comprising 80 real-world tables, with splits of 49/16/15 for training, development, and testing for TQA task. It contains five types of question-answer pair: boolean, category, number, list[category], and list[number]. The test set contains 522 human-annotated question-answer pairs. We categorize the tables of test set into large, medium, and small sets based on the number of cells. The distribution of QA types and the division of size are detailed in Appendix \ref{sec:appendix_databench}. DataBench offers a reduced version called DataBench Lite, where each table retains the first 20 rows of data. Correspondingly, the competition includes two subtasks: subtask A on DataBench and subtask B on DataBench Lite.


\subsection{Related works}

TableLlama \citep{zhang-etal-2024-tablellama}, TableGPT \citep{zha2023tablegptunifyingtablesnature}, and StructLM \citep{zhuang2024structlm} continue to pre-train models with the decoder-only architecture like Llama\citep{touvron2023llamaopenefficientfoundation} towards various table-related tasks. However, table tuning demands a substantial amount of high-quality labeled data, which may be difficult to obtain in certain domains.


With the progress of LLMs, the paradigm has increasingly shifted from traditional models pre-trained from scratch or task-specific modules designed for narrow applications \citep{LIU2023127,xiong-etal-2024-dual} to leveraging LLMs' strong reasoning and emergent abilities for adapting to downstream tasks \citep{ruan2024language,mrsql-zhenhe,Zhongqiu2024,10884554}. LEVER \citep{ni2023lever}, Binder \citep{Binder}, PoTable \citep{mao2024potableprogrammingstandardlytablebased} and OpenTab \citep{kong2024opentab} are programming-based methods, solving questions with the assistance of SQL or Python codes. However, these methods exhibit limitations in fully comprehending the relationships between questions and table data. Dater \citep{dater}, Chain-of-Table \citep{wang2024chain} and ReActTable \citep{ReAcTable} prompt LLMs iteratively to locate critical rows and columns. These methods feed the entire table text into the model, which are usually not applicable for larger tables due to inherent context length constraint. TableRAG \citep{NEURIPS2024_tablerag} introduces a million-token table understanding framework, while the encoding and matching operations result in accuracy loss and additional budget.



\section{System Overview}
\begin{figure*}[htbp]
    \centering 
    \includegraphics[width=0.95\textwidth]{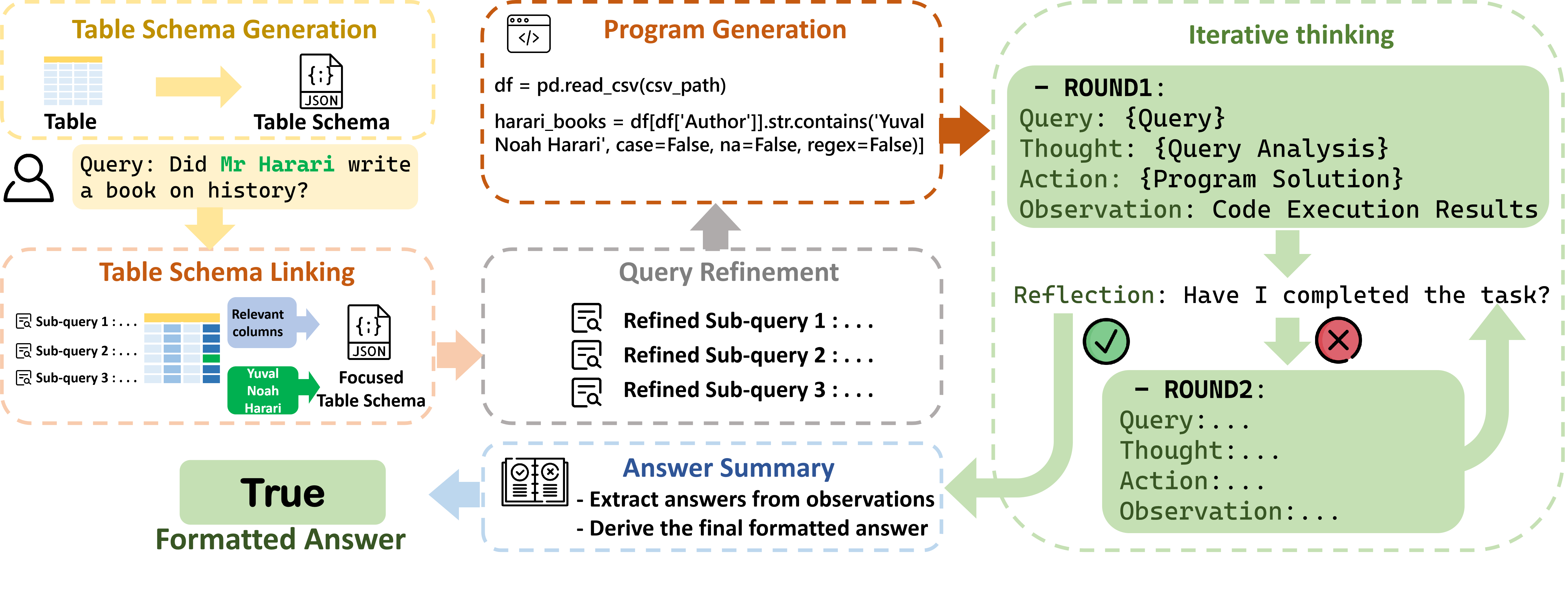} 
    \caption{Architecture of TableReasoner framework. The table is firstly converted into a table schema containing global features and example values. The table schema and query go through a reasoning workflow which comprising 5 core sequential modules to obtain observations. The reasoning process and observations are then fed into an iterative thinking framework. It's repeated until expectations are met.}
    \label{fig:workflow}
\end{figure*}

Our system is implemented using TableReasoner, an LLM-powered and programming-based framework, as shown in Figure \ref{fig:workflow}. 
It's designed to integrate a reasoning workflow into an iterative thinking process.


 


\subsection{Reasoning Workflow}

\firstsentencebold{Table Schema Generation.}\quad
We utilize the table schema to describe and provide table information to LLMs.
Firstly, we use Python to read the spreadsheet file. Each column is considered as a distinct feature, characterized by data type and meta statistical attributes (e.g., the maximum, minimum, mean and median values of numerical data; unique categories and the most frequently occurring items of categorical data). Additionally, $K$ random rows are selected as example values. Then, to disambiguate specific column names (such as abbreviations), we incorporate the latent semantics of column names into the table schema. In detail, we treat the above table metadata as a preliminary schema to prompt the LLM to generate descriptions of the table and each column. The global table schema is structured as JSON format, an illustrative example is provided in Appendix \ref{sec:appendix_schema}. 

This table representation method expands the capacity of TableReasoner in processing large-sized tables. The token complexity of a table schema is approximately \(O(N)\) , which is significantly lower than the complexity \(O(M\times N)\) of the entire table when \(M\) is very large, where \(M\) and \(N\) denote the number of rows and columns respectively.

\firstsentencebold{Table Schema Linking.}\quad
We propose a "parsing-linking-refinement" action flow to refine the global table schema into a focused schema strongly associated with the query. 
 (i) Parsing. We prompt LLM to parse the query based on the table schema, decoupling it into specific sequential sub-queries. (ii) Linking. During parsing, the LLM is also prompted to extract relevant columns for each sub-query, a process referred to as \textit{column linking}. To align the entities mentioned in the query with the values in the table, an \textit{entity linking} step is employed. Specifically, the LLM is used to identify whether the query contains entities, extract them, and suggest belonging table columns. Next, Python is utilized to read all elements of the corresponding columns, and the Longest Common Subsequence algorithm is applied to retrieve elements from column entries whose overlap rate with the query entity exceeds 0.6. Finally, we use the LLM to precisely select the aligned value from these recalled elements. For instance, "\textit{Mr Harari}" in the query is linked to "\textit{Yuval Noah Harari}" in the table. (iii) Refinement. At last, we prune irrelevant columns from the global table schema and integrate entity alignment information as needed, yielding a focused table schema. This module effectively reduces token consumption and noise input.

\firstsentencebold{Query Refinement.}\quad
To enhance comprehension of complex reasoning tasks, we further decompose the query into $S$ progressive sub-queries along with associated column names using Chain-of-Thought (CoT) prompting \citep{wei2022chain}. The sole distinction from query parsing in table schema linking stage lies in employing the focused table schema which is more information-dense and de-noised.

\firstsentencebold{Program-assisted Solution Generation.}\quad
The TableReasoner framework centers on programming to derive precise results. Specifically, we guide the LLM to generate a Program-of-Thoughts (PoT) \citep{chen2023program} solution, utilizing the focused table schema and refined queries from prior steps. The generated codes are subsequently executed in isolated environments (e.g., Python interpreter) to produce verifiable results. Compared to only textual reasoning approach, the program-assisted solution can effectively mitigate numerical hallucinations in multi-step data acquisition and processing.

\firstsentencebold{Answer Summary.}\quad
Due to the strict answer format requirements of DataBench, we introduce an answer summary module at the end of the workflow. 
This module generates the final formatted answer by summarizing intermediate thoughts and observations derived from Python executions during the iterative reasoning process.

\subsection{Iterative Thinking Paradigm}\quad 
Inspired by the ReAct, we design the iterative thinking paradigm, integrating the reasoning workflow into a "\textit{thought-action-observation}" architecture, with the goal of bringing incremental self-reflection and decision-making mechanisms into the framework. "\textit{Thought}" corresponds to the decomposed sub-queries from the Query Refinement stage; "\textit{action}" represents the program ideas and codes generated during the Program-assisted Solution Generation stage; and "\textit{observation}" is the feedback from code execution. This creates a reasoning cycle in our workflow. Upon completion of each cycle, the system evaluates whether the query can be answered with the current reasoning state. If yes, it proceeds to the Answer Summary stage; otherwise, it generates a new follow-up query and repeats the process.

\subsection{Supervised Fine-tuning}
For the purpose of boosting  the QA ability on DataBench dataset, we  fine-tune the LLMs deployed in Query Refinement and Program-assisted Solution Generation stages. We adopt rejection sampling \citep{yuan2023scaling} method on DataBench train and development datasets to synthetize training data. Please refer to Appendix \ref{sec:appendix_finetune} for details.

\section{Experimental setup}
\label{sec:exp_setup}

\begin{table*}[h!]
\centering
\small
\begin{tabular}{cc|ccccc}
\toprule
\textbf{Model} & \textbf{Avg} & \textbf{boolean} & \textbf{category} & \textbf{number} & \textbf{list[category]} & \textbf{list[number]} \\
\midrule
\textbf{Code-based} & & & & & & \\
Qwen2.5-7B         & 69.15 / 69.92 & 73.64 / 76.74 & 63.74 / 61.54 & 70.51 / 69.87 & 59.72 / 65.28 & 72.97 / 74.32 \\
Qwen2.5-32B       & 77.39 / 77.59 & 85.27 / 89.15 & 68.13 / 72.53 & 70.51 / 72.44 & 83.33 / 75.00 & 82.43 / 78.38 \\
Qwen2.5-32B-Coder  & 81.03 / 81.03 & 93.02 / 89.15 & 78.02 / 74.73 & 76.28 / 79.49 & 73.61 / 77.78 & 79.73 / 82.43 \\
Qwen2.5-72B        & 81.03 / 81.22 & 86.05 / 89.92 & 80.22 / 71.43 & 82.69 / 82.05 & 77.78 / 76.39 & 71.62 / 81.08 \\
Llama3.1-8B        & 51.15 / 51.72 & 55.81 / 50.39 & 36.26 / 39.56 & 66.03 / 66.67 & 43.06 / 37.50 & 36.49 / 52.70 \\
Llama3.3-70B       & 74.14 / 77.39 & 79.84 / 88.37 & 76.92 / 68.13 & 73.72 / 78.21 & 72.22 / 69.44 & 62.16 / 77.03 \\
TeleChat2-35B      & 71.07 / 78.54 & 86.82 / 86.82 & 73.63 / 73.63 & 62.82 / 76.92 & 68.06 / 73.61 & 59.46 / 79.73 \\
Mistral-Large & 85.63 / 83.91 & \underline{95.35} / 93.02 & 78.02 / 81.32 & 83.97 / 83.33 & 83.33 / 73.61 & 82.43 / 83.78 \\
GPT-4o              & 85.44 / 83.72 & \textbf{96.90} / 93.02 & 76.92 / 76.92 & 83.97 / 83.33 & 79.17 / 75.00 & 83.78 / 86.49 \\
\midrule
\textbf{TableReasoner without SFT} & & & & & & \\
Qwen2.5-7B   & 81.61 / 82.95 & 87.6 / 90.6 & 74.73 / 76.92 & 79.49 / 82.05 & 84.72 / 80.41 & 81.08 / 86.49 \\
Qwen2.5-32B  & 89.85 / \underline{89.66} & \underline{95.35} / \textbf{95.35} & \underline{86.81} / \textbf{87.91} & \underline{86.54} / 85.26 & \textbf{89.27} / \textbf{87.44} & 89.19 / \underline{94.59} \\
Qwen2.5-72B      & 87.55 / 88.31 & 92.25 / \textbf{95.35} & 80.22 / 84.62 & \underline{86.54} / 85.90 & \underline{86.11} / \underline{84.67} & \underline{90.54} / 89.19 \\
Mistral-Large     & \underline{90.23} / \textbf{90.04} & \underline{95.35} / 93.02 & \textbf{90.11} / \textbf{87.91} & \textbf{88.46} / \textbf{89.74} & 84.72 / 80.50 & \underline{90.54} / \textbf{97.30} \\
Combination$^\triangle$ & \textbf{90.61} / \textbf{90.04} & \underline{95.35} / \underline{94.57} & \textbf{90.11} / \underline{86.81} & \underline{86.54} / \underline{88.46} & \underline{86.11} / \underline{84.67} & \textbf{94.59} / \underline{94.59} \\
\bottomrule
\end{tabular}
\caption{Performance comparison of Accuracy(\%) on DataBench / DataBench Lite test sets. $\triangle$ means the combination of Mistral-Large for Program-assisted Solution Generation and Qwen2.5-32B for other modules.}
\label{table:exp_main_res}
\end{table*}


\firstsentencebold{Model}\footnote{Unless stated otherwise, all models employed in the experiments are the Instruct versions.}.\quad We conduct extensive experiments on popular LLMs, including the open-sourced Qwen2.5 series \citep{qwen2.5, hui2024qwen2}, Llama3 series \citep{llama3}, TeleChat2-35B \citep{he2024telechattechnicalreport,li2024teleflmtechnicalreport}, Mistral-Large\footnote{\url{https://huggingface.co/mistralai/Mistral-Large-Instruct-2407}} and close-sourced GPT-4o \citep{gpt4}. GPT-4o is invoked via the official API interface, and other models are deployed and invoked locally. 

\firstsentencebold{Implementation}.\quad We fine-tune the Qwen2.5-32B-Instruct and Mistral-Large models for the Query Refinement and Program-assisted Solution Generation stages respectively, applying Low-Rank Adaptation method \citep{hu2022lora}. Each model is trained for 5 epochs, with learning rate of 5e-6, lora rank of 8 and total batch size of 32. We set the temperature to 0 to ensure stable output during inference. If majority-voting strategy is adopted, the temperature is configured according to the default values. Within the iterative thinking framework, the maximum round of reasoning cycle is set to 5. We design delicate prompts combining few-shot learning, structured output and CoT strategies to mitigate model bias and address intricate details. Some prompts can be found in Appendix \ref{sec:appendix_prompts}.

\firstsentencebold{Baseline}.\quad We compare 2 common prompting approaches with our TableReasoner. \textbf{(i) Zero-shot In-Context Learning} (Z-ICL), which provides the task description and table data in the prompt for textual reasoning. \textbf{(ii) Code-based}, which directly answers queries by generating PoT solution leveraging a single LLM and executing codes. For the sake of fairness, we format program execution results through the Answer Summary module to produce final answers. 
In these cases, we represent tables in Markdown format, which is one of the most common verbalized types of tabular data.

\section{Results and Analysis}
\label{sec:results}

\subsection{Main Results}
The main results are shown in Table \ref{table:exp_main_res}. For the code-based approach, Mistral-Large and GPT-4o show superior performance. A consistent trend of performance enhancement is observed as the model parameters increased in Qwen model series. The results of the Z-ICL method are provided in Appendix \ref{sec:appendix_zicl}.

\begin{figure}[ht!]
\setlength{\abovecaptionskip}{3pt}
\setlength{\belowcaptionskip}{3pt}
    \begin{minipage}{\linewidth}
    \centering
     \includegraphics[width=1.05\linewidth]{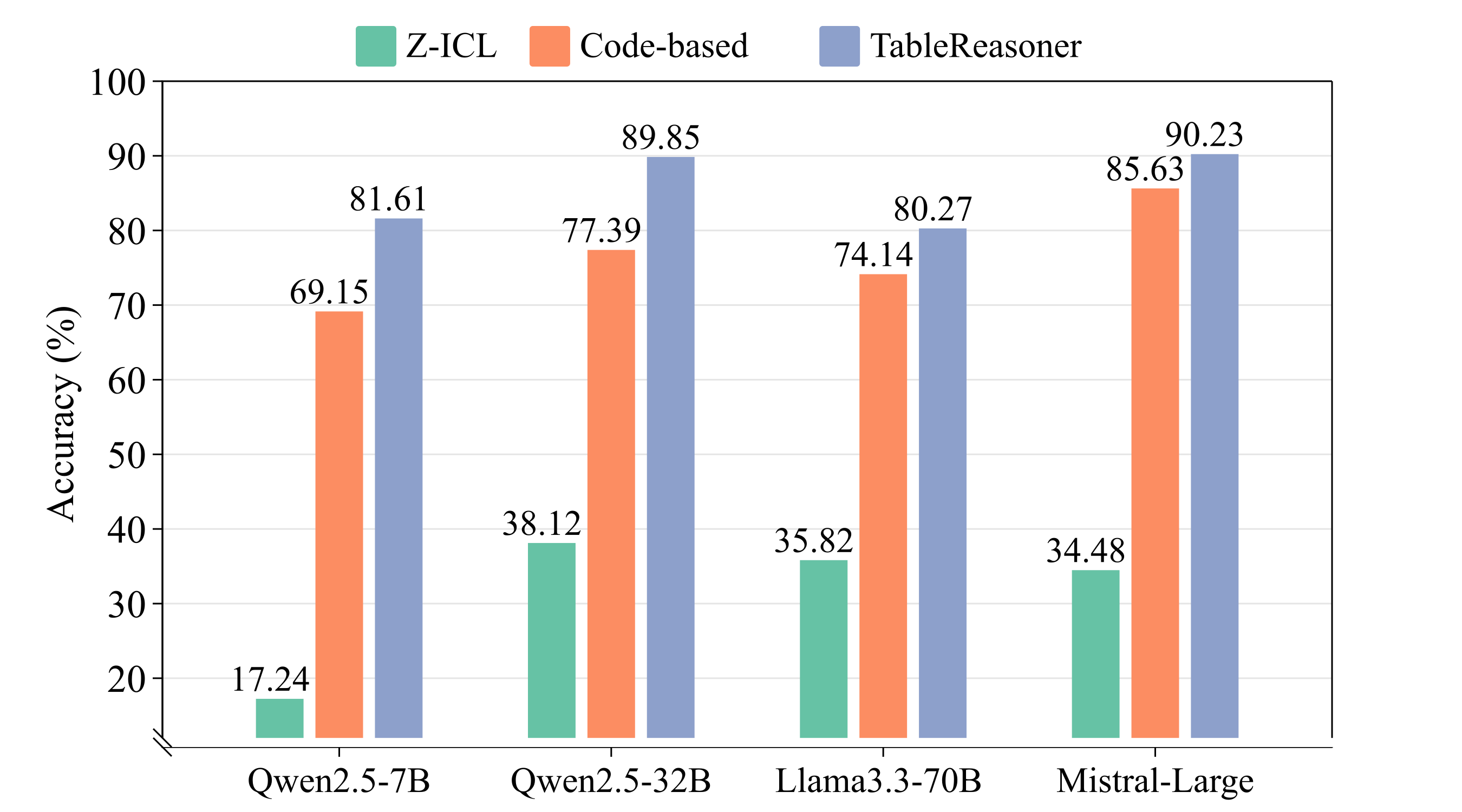}
    \caption{Accuracy comparison with baselines on DataBench test set.}
    \label{fig:method_compare_on_test}
    \end{minipage}
    \vfill
    \begin{minipage}{\linewidth}
    \centering
    \includegraphics[width=1.05\linewidth]{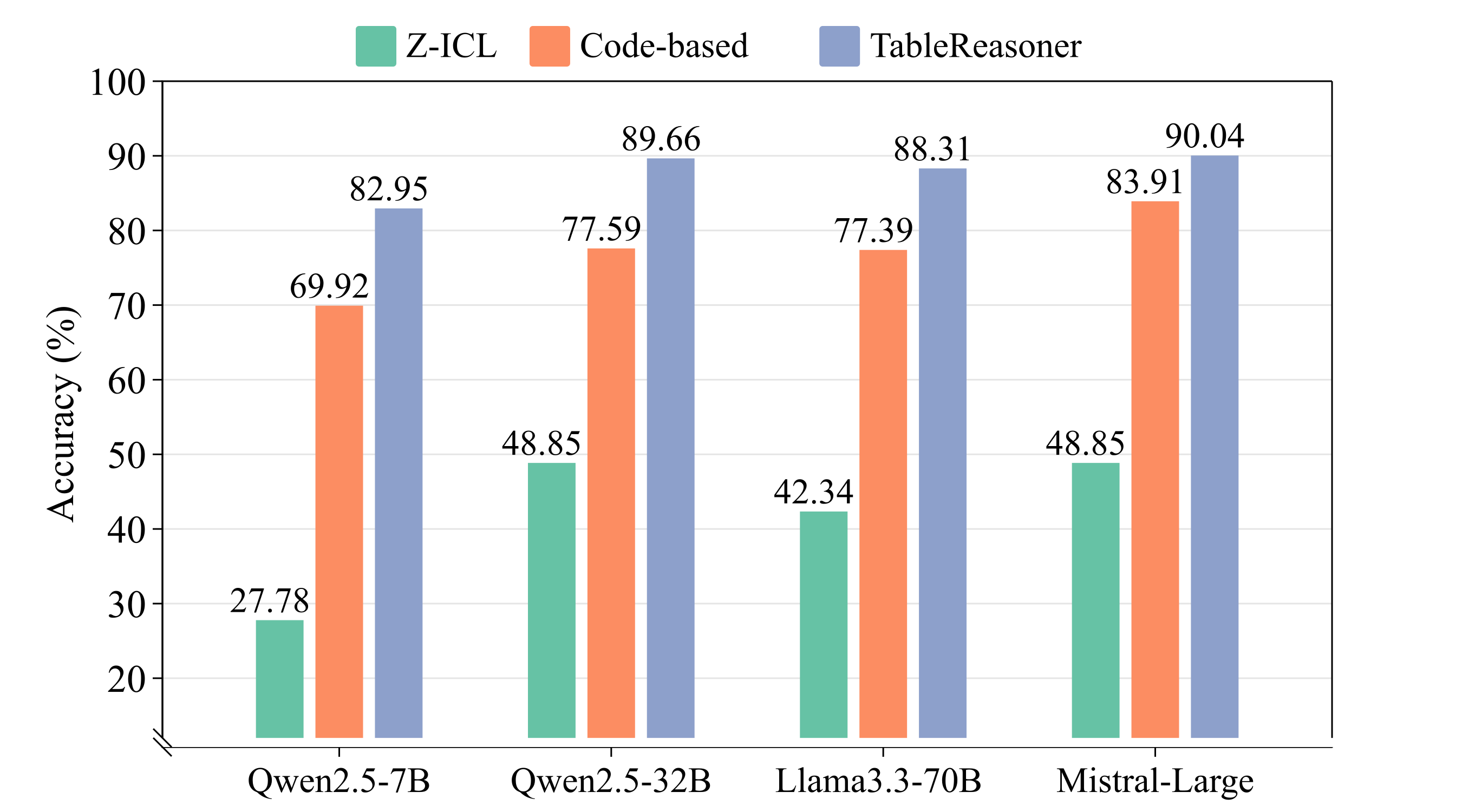}
    \caption{Accuracy comparison with baselines on DataBench Lite test set.}
    \label{fig:method_compare_on_testlite}
    \end{minipage}
\end{figure}

\begin{figure}[t]
	\centering 
	\includegraphics[width=1\linewidth]{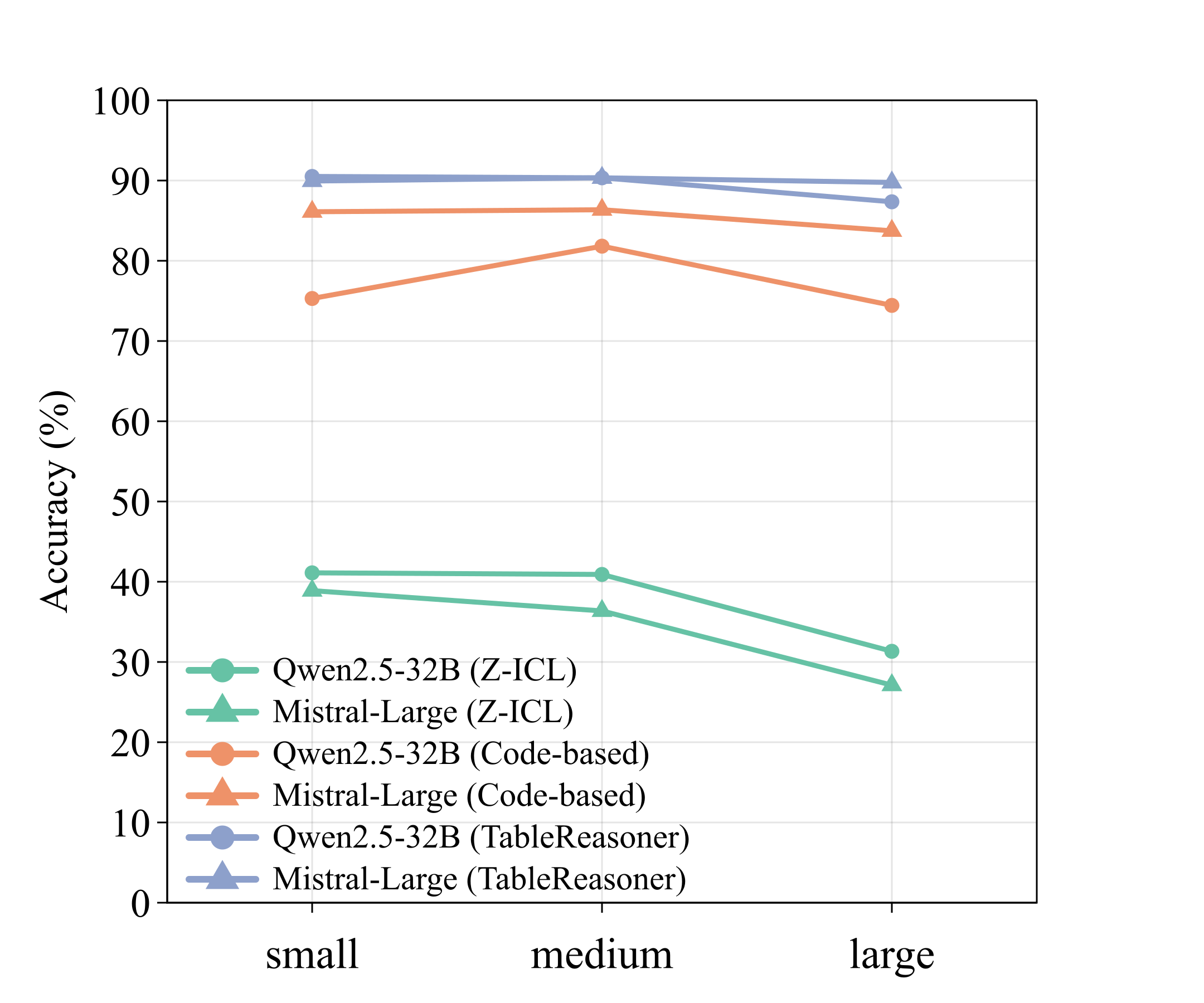}
	\caption{Accuracy comparison between tables of different sizes on DataBench test set.} 
	\label{fig:size_compare}
\end{figure}

\firstsentencebold{Improvements brought by TableReasoner}. TableReasoner consistently enhances the performance across both DataBench and DataBench Lite without fine-tuning. As evidenced in Figure \ref{fig:method_compare_on_test} and \ref{fig:method_compare_on_testlite}, it brings remarkable improvements in accuracy, with absolute increases of 40\%+ under the configurations of various backbone models, when compared to the Z-ICL approach. It also realizes substantial improvements to the code-based approach. Moreover, TableReasoner narrows the performance gap between models of small and large parameter size. Notably, integrating Qwen2.5-7B into TableReasoner achieves high accuracy scores, surpassing the top performer in the small-scale model competition rankings (fewer than 8B parameters).

\firstsentencebold{Analysis on different QA types}. Table \ref{table:exp_main_res} shows that models perform well on Boolean but struggle with list[category]  QA types. These complex queries require the model to have a profound understanding of user intent and possess a deep comprehension of tabular data, including deduplication and multi-column correlation analysis. The proposed TableReasoner exhibits more balanced performance advantages across various QA types.

\firstsentencebold{Analysis on table size}. We explore the performance on small, medium and large size of tables. As depicted in Figure \ref{fig:size_compare}, the Z-ICL approach experiences a sharp performance decline as the table size increases. The code-based approach consistently outperforms Z-ICL and maintain relatively stable performance on tables of different sizes. It not only validates the efficacy of the PoT method but also highlights its great potential for handling large-scale tabular data. Remarkably, TableReasoner exhibits outstanding scalability and robustness, showing minimal degradation as the table size expands.


\subsection{Ablation Study}
We perform ablation study to verify the effectiveness of each component of TableReasoner workflow. The experiments are conducted on the framework that embeds Qwen2.5-32B model, and results are shown in Table \ref{tab:ablation}. The accuracy shows a drastic deterioration when removing table schema generation and schema linking modules and replacing table schema with the markdown-format text in other remaining stages. The accuracy decreases by 5.37\% on the test set and 1.92\% on the Lite test set, strongly demonstrating the superiority of applying table schema as the representation, especially for large-sized tables. It is suggested that the specific data in each row is of lesser importance compared to comprehending table's overall structure and column features for programming-based solutions. Removing the schema linking or query refinement module leads to a drop in accuracy, highlighting the importance of the focused table schema and refined sub-queries. 


\begin{table}[h] 
\centering
\begin{tabular}{lll} 
\toprule
\textbf{Method} & \textbf{Test} & \textbf{Lite Test} \\
\midrule
TableReasoner & 89.85 & 89.66 \\
\ - table schema & 84.48 \scriptsize{($\downarrow$ 5.37)} & 87.74 \scriptsize{($\downarrow$ 1.92)} \\
\ - schema linking & 87.55 \scriptsize{($\downarrow$ 2.30)} & 88.31 \scriptsize{($\downarrow$ 1.35)} \\
\ - query refinement & 88.51 \scriptsize{($\downarrow$ 1.34)} & 89.27 \scriptsize{($\downarrow$ 0.39)} \\
\bottomrule 
\end{tabular}
\caption{Ablation results on DataBench and DataBench Lite test sets.}  
\label{tab:ablation} 
\end{table}

\vspace{-5pt} 
\begin{table}[h] 
\centering
\begin{tabular}{lcc} 
\toprule
\textbf{Method} & \textbf{Test} & \textbf{Lite Test} \\
\midrule
TableReasoner & 90.61 & 90.04 \\
\ \ +\ Fine-tuning & 92.53 & 91.19 \\
\ \ +\ Majority-voting (k=5) & 93.87 & 91.76 \\
\bottomrule 
\end{tabular}
\caption{Results of useful strategies.} 
\label{tab:strategy}
\end{table}

\vspace{-5pt} 

\subsection{Effects of Strategies}
\firstsentencebold{LLM Combination}. We observe Mistral performs better in code generation tasks. Building upon this finding, we design a hybrid architecture: employing Mistral-Large in Program-assisted Solution Generation stage, while retaining Qwen2.5-32B for other functional modules. The experimental results shown in Table \ref{table:exp_main_res} indicate that this hybrid architecture slightly improves accuracy by 0.76\% on DataBench test set compared to the architecture using solely Qwen2.5-32B.

\firstsentencebold{Fine-tuning \& Majority-voting}. To further improve the accuracy on DataBench, we employ the fine-tuned LLMs described in Section \ref{sec:exp_setup} within the hybrid architecture. As demonstrated in Table \ref{tab:strategy}, coupled with majority voting based on the self-consistency principle, our system achieves state-of-the-art accuracy of 93.87\% and 91.76\% on test and Lite test sets, respectively.

\section{Conclusion}
In this paper, we introduce an LLM-powered and programming-based table easoning framework — TableReasoner, which achieves first place in both subtasks of SemEval-2025 Task 8. We utilize the table schema as a representation, understanding the table from a holistic perspective and addressing the context length constraint. A comprehensive schema linking is implemented in TableReasoner, which provides a focused and precise table schema for query refinement and programming. Besides, we propose an iterative thinking paradigm, incorporating the reasoning workflow to facilitate incremental thinking and reflection. We further enhance performance through fine-tuning and majority voting. Extensive experiments indicate our system is of high scalability and performance across real-world table datasets and has a greater advantage on large-sized tables. 


\section*{Limitations}
For Z-ICL and Code-based methods using LLMs without fine-tuning, prompt design plays a critical role in influencing model performance. For instance, representing tabular data in different formats, such as JSON, CSV or Markdown, can lead to varying results, as different LLMs may exhibit format preferences. Due to time constraints, we were unable to comprehensively evaluate the effects of prompt variations on the comparative experimental outcomes. We will further investigate the impact of different prompt designs.

In this study, we focus on the domain of reasoning over tabular data. Our proposed TableReasoner framework achieves high accuracy by encouraging deliberate reasoning steps. However, it requires numerous inference iterations which are time-consuming. In future work, We will explore adaptive action flow to balance inference times with accuracy.

\bibliography{custom}

\appendix

\section{Distribution of DataBench Test Set}
\label{sec:appendix_databench}

\begin{figure}[h]
    \flushleft 
    \includegraphics[width=0.4\textwidth]{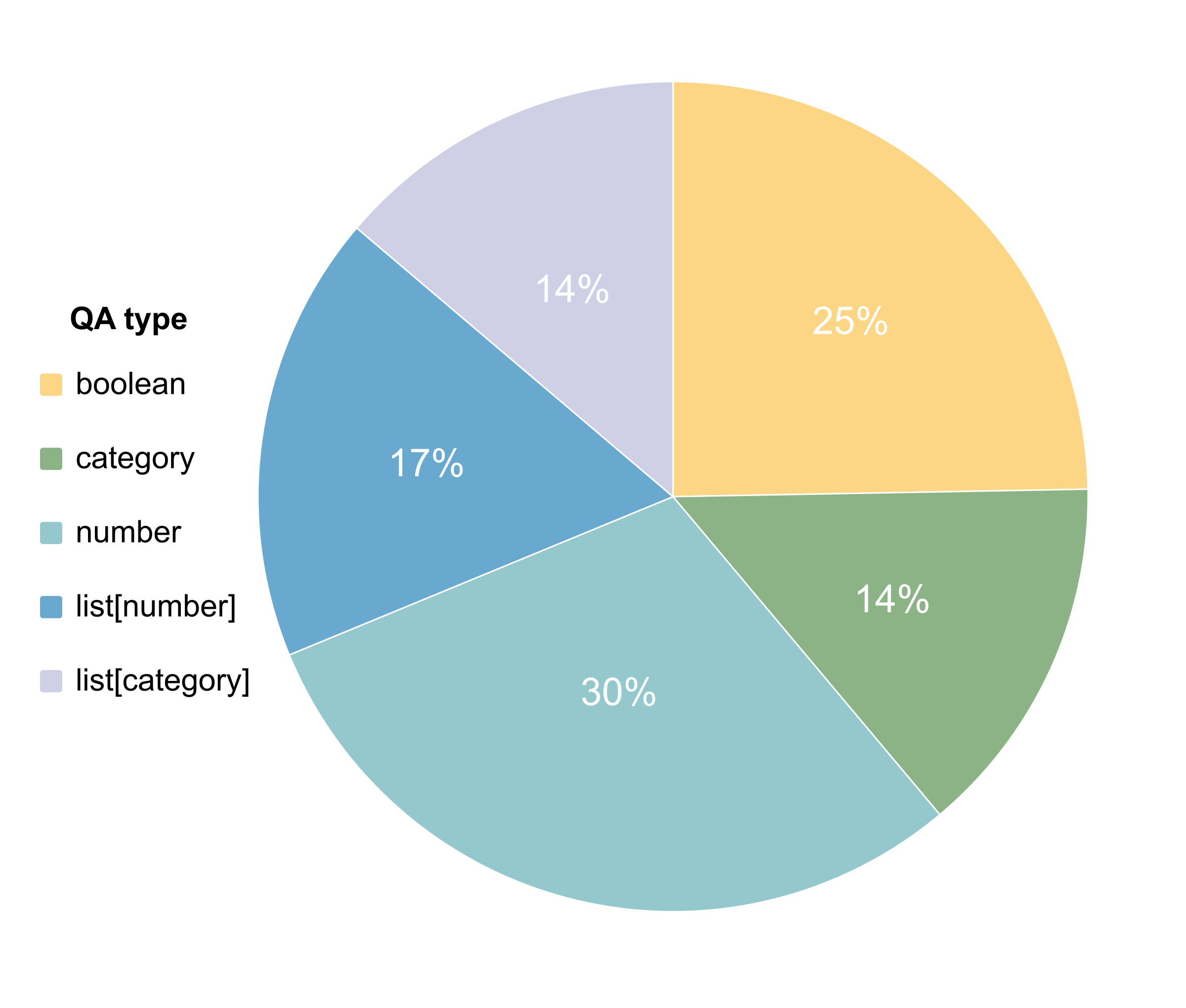} 
    \caption{The distribution of Question-Answer pair types on test set.}
    \label{Fig:background_pie}
\end{figure}

The distribution of QA-pair types is depicted in Figure \ref{Fig:background_pie}. We split tables in DataBench test set into large, medium, and small groups based on the number of cells, as shown in Table \ref{tab:size_split}. The small tables include 069\_Taxonomy, 071\_COL, 072\_Admissions, 075\_Mortality and 080\_Books, total 180 questions; the medium tables include 066\_IBM\_HR, 073\_Med\_Cost, 074\_Lift, 077\_Gestational and 078\_Fires, total 176 questions; and the large tables include 067\_TripAdvisor, 068\_WorldBank\_Awards, 070\_OpenFoodFacts, 076\_NBA and 079\_Coffee, total 166 questions.

\begin{table}[!htbp]
    \centering
    \begin{tabular}{ccc}
    \toprule 
        \textbf{Type} & \textbf{Dataset Number} & \textbf{Average Cells}\\
        \hline
        Large & 5 & 1519065 \\
        \hline
        Medium & 5 & 18131 \\
        \hline
        Small & 5 & 2882 \\
    \bottomrule
    \end{tabular}
    \caption{The split of table size on DataBench test set.} 
    \label{tab:size_split}
\end{table}

\begin{table*}[!ht]
\centering
\small
\begin{tabular}{ccccccc}
\toprule
\textbf{Model Name} & \textbf{Avg} & \textbf{boolean} & \textbf{category} & \textbf{number} & \textbf{list[category]} & \textbf{list[number]} \\
\midrule
Qwen2.5-7B         & 17.24 / 27.78 & 41.86 / 53.49 & 3.30 / 13.19  & 9.62 / 21.79  & 5.56 / 11.11  & 17.57 / 31.08 \\
Qwen2.5-32B       & 38.12 / 48.85 & 82.95 / 75.19 & 26.37 / 48.35 & 20.51 / 39.74 & 16.67 / 27.78 & 31.08 / 44.59 \\
Qwen2.5-32B-Coder  & 32.37 / 41.95 & 76.74 / 73.64 & 16.48 / 37.36 & 17.95 / 32.05 & 13.89 / 23.61 & 21.62 / 32.43 \\
Qwen2.5-72B        & 41.95 / 51.92 & 76.74 / 86.05 & 26.37 / 51.65 & 30.13 / 39.74 & 20.83 / 23.61 & 44.59 / 47.30 \\
Llama3.1-8B        &  29.31 / 25.10  & 62.79 / 45.74  & 21.62 / 25.68  &    17.31 / 22.44 & 12.50 / 12.50 & 21.98 / 9.89 \\
Llama3.3-70B       & 35.82 / 42.34 & 79.84 / 84.50 & 18.68 / 21.98 & 17.31 / 27.56 & 23.61 / 25.00 & 29.73 / 43.24 \\
TeleChat2-35B      & 32.18 / 42.72 & 75.97 / 75.97 & 14.29 / 36.26 & 18.59 / 32.69 & 15.28 / 25.00 & 21.62 / 32.43 \\
Mistral-Large & 34.48 / 48.85 & 71.32 / 71.32 & 25.27 / 51.65 & 19.23 / 42.31 & 18.06 / 29.17 & 28.38 / 40.54 \\
GPT-4o              & 41.57 / 56.32 & 84.50 / 83.72 & 28.57 / 59.34 & 26.92 / 52.56 & 20.83 / 25.00 & 32.43 / 44.59 \\
\bottomrule
\end{tabular}
\caption{Performance of Z-ICL prompting approach on DataBench / DataBench Lite test sets. The metric is accuracy(\%).}
\label{tab:z-icl_res}
\end{table*}

\section{Table Schema Example}
\label{sec:appendix_schema}
\definecolor{codegreen}{rgb}{0,0.6,0}
\definecolor{codegray}{rgb}{0.5,0.5,0.5}
\definecolor{codepurple}{rgb}{0.58,0,0.82}
\definecolor{backcolour}{rgb}{0.95,0.95,0.92}

\lstdefinestyle{json}{
    backgroundcolor=\color{backcolour},
    commentstyle=\color{codegreen},
    keywordstyle=\color{magenta},
    numberstyle=\tiny\color{codegray},
    basicstyle=\ttfamily\footnotesize,
    numbers=none,    
    breakatwhitespace=false,
    breaklines=true,
    captionpos=b,
    keepspaces=true,
    numbersep=5pt,
    showspaces=false,
    showstringspaces=false,
    showtabs=false,
    tabsize=2
}

\begin{lstlisting}[style=json, label={lst:json-example},linewidth=0.5\textwidth]
{
    "file_path": "072_Admissions/all.csv",
    "table_name": "072_Admissions",
    "table_description": "The Admissions table contains information about applicants for graduate programs, including their test scores, academic performance, and the likelihood of being admitted. This data can be used to analyze factors that influence admission chances and to predict admission outcomes for new applicants.",
    "number_of_rows": 500,
    "column_list": [
        "Serial No.",
        "GRE Score",
        "TOEFL Score",
        "University Rating",
        "SOP",
        "LOR",
        "CGPA",
        "Research",
        "Chance of Admit"
    ],
    "column_description": [
            ...(omit)...
        {
            "column_name": "SOP",
            "dtype": "float64",
            "example": {
                "minimum_value": 1.0,
                "maximum_value": 5.0,
                "median_value": 3.5,
                "average_value": 3.374
            },
            "specific_meaning": "The Statement of Purpose (SOP) score of the applicant, on a scale of 1 to 5."
        },
        ...(omit)...
    ]
    "cell_example": [
        {
            "Serial No.": 469,
            "GRE Score": 323,
            "TOEFL Score": 110,
            "University Rating": 4,
            "SOP": 4.0,
            "LOR": 5.0,
            "CGPA": 8.88,
            "Research": 1,
            "Chance of Admit": 0.81
        },
        ...(omit)...
        ]
} 
\end{lstlisting}

\section{Preparation of Fine-tuning Dataset}
\label{sec:appendix_finetune}
The preparation of our fine-tuning dataset involves the following steps:  
(1) \textbf{Reasoning Path Collection}. Within our pipeline, we utilize multiple advanced LLMs—such as GPT-4o, Qwen2.5-72B-Instruct, and Mistral-Large—to perform five rounds of inference on both the training and development sets of DataBench (excluding the DataBench Lite subset), with the temperature parameter set to its default value (not 0). The inputs and outputs of the Query Refinement and Program-assisted Solution Generation modules are recorded for each inference round. Consequently, for each question \( q \), we obtain two sets of reasoning paths: \( G[q, (u_i, u_o), (c_i, c_o)] \), where $u_i$ and $u_o$ denote the input and output of Query Refinement step, $c_i$ and $c_o$ denote the input and output of Program-assisted Solution Generation step. (2) \textbf{Filtering Incorrect Reasoning Paths}. We verify the correctness of the final reasoning results against ground truth answers and discard any reasoning paths that lead to incorrect results.(3) \textbf{Reasoning Path Selection}. We employ a rule-based reward mechanism to select the optimal reasoning paths. For the Query Refinement model, we prioritize paths  \( (u_i, u_o) \) that correctly associate column names with more sub-queries. For the Program-assisted Solution Generation model, we select paths  \( (c_i, c_o) \) with greater code length. If multiple candidates remain, one is chosen randomly.

For questions that have never been answered correctly, we manually remove ambiguous ones and annotate the others with hints. Specifically, we provide one to three key clues to guide the LLMs toward correctly interpreting and answering the question. The question and its corresponding hints are then concatenated and reprocessed following the aforementioned steps.

After filtering and annotating as described above, the fine-tuning dataset consists of 1184 out of 1308 training samples,

\section{Performance of Z-ICL approach}
\label{sec:appendix_zicl}
As shown in Table \ref{tab:z-icl_res}, the performance of Z-ICL approach falls short on both DataBench test and Lite test sets, with the highest scores being 41.38/56.70, achieved by leveraging GPT-4o. This Z-ICL paradigm confronts several intractable challenges, including text truncation due to length limitations, potential hallucination phenomena, and the absence of explicit reasoning processes.

\onecolumn
\section{Prompts}
\small
\label{sec:appendix_prompts}







\subsection{Table Description Prompt}

\begin{tcolorbox}[sidebyside, sidebyside align=top seam, width=\linewidth, colback=gray!20, colframe=white, colbacktitle=white, coltitle=white, breakable, arc=0mm, left=0mm, right=0mm]

Given a database schema regarding "\{table\_name\}", your task is to analyse all columns in the database and add detailed explanations for database and each column.\\

Requirements:\\
1.Response should include column names and the specific meanings of each column to help users better understand the data content.\\
2.Response format example:\\
\{

    \hspace{15pt}"Table\_Description":..., 
    
    \hspace{15pt}"Column\_Description": [
    
        \hspace{30pt}\{"column\_name": "Age", "specific\_meaning": "Represents User's Age."\},
        
        \hspace{30pt}\{"column\_name": "Joined Date", "specific\_meaning": "The date on which the user joined."\},
        
        \hspace{30pt}\{"column\_name": "Gender", "specific\_meaning": "User's Gender, with 2 categories."\},
        
        \hspace{30pt}\{"column\_name": "City", "specific\_meaning": "City where the user resides, with 32 categories and only one category example displayed."\}]
        
\}\\

Definition of fields:\\
**Table\_Description**: Explain the main content and possible uses of the table.\\
**Column\_Description**: Explain the meaning of each column.\\
Ensure that the response format is a compact and valid JSON format without any additional explanations, escape characters, line breaks, or backslashes.\\

\#\#\# Database Schema\\
\{table\_schema\}\\

Please response in **JSON** format complying with the above requirements.

\end{tcolorbox}

\subsection{Query Refinement Prompt}

\begin{tcolorbox}[sidebyside, sidebyside align=top seam, width=\linewidth, colback=gray!20, colframe=white, colbacktitle=white, coltitle=white, breakable, arc=0mm, left=0mm, right=0mm]

As an experienced and professional data analysis assistant, your goal is to analyze a user's question and identify the relevant columns that might contain the necessary data to answer user's question based on the table schema.  The table schema consists of table descriptions and multiple column descriptions. \\

Specifically, you need to complete two sub tasks: \\
\texttt{[task1]}\\

Thoroughly understand and analyze user's question. You should orient your approach towards resolving user query by referencing the information provided in the table schema, and break down the original query into more specific, complete and executable sub-queries.\\
\texttt{[task2]}\\

For each query to be answered, identify and extract the relevant columns from the `column\_list` field in the table schema that are necessary to answer the query.\\

\#\#\# Instruction\\
\texttt{[task1 Instruction]}\\
- You should attempt to decompose the original query into more specific, progressively detailed, step-by-step sub-queries. Ensure the sub-queries maintain high relevance to the original query and executability to table retrieval, and confirm that no critical information is omitted.\\
- You can recognize key entities, intentions, special reminder, and specific objects from user's question, which can help you accurately analyze user issues.\\
- Ensure that each query can be answered by retrieving relevant values from the table.\\
- Pay attention to the expression of the maximum value (maximum/top/highest/most/lowest/smallest/last, etc) in user's query.\\

\texttt{[task2 Instruction]}\\
- Identify one or more relevant columns from the `column\_list` field in the table schema that are necessary to answer each query. \\
- Distinguish between easily confused column names, and refer to column descriptions and example values if necessary, to ensure the accuracy of the relevant columns extracted.\\
- The user's terminology may have multiple meanings or their expression might be ambiguous. In such cases, try to infer the most likely intent from the user's query and provide all potentially relevant columns.\\
- When queries are vague or ambiguous, attempt to infer the most likely intent based on the user's question and the table description, and provide all potentially relevant columns as comprehensively as possible.\\
- Ensure no necessary columns are omitted.\\
- Please reflect and ensure that the extracted column names must exist in the table schema(`Field: column\_list`). Prohibit modification and avoid any illusions to ensure that the relevant values can be read from the table. \\

\#\#\# Output format\\
Please answer with a list of sub\_queries in JSON format **without any additional explanation**.\\

Examples:\\
**Question**: What are the average sales, cost, and profit per order for children's food?\\
**Response**:[

\hspace{15pt}\{"Query1": "Filter the data to include only orders related to children's food.", "relevant\_column\_list": ["product\_category"]\},

\hspace{15pt}\{"Query2": "Calculate the average sales per order for children's food.", "relevant\_column\_list": ["sales"]\},

\hspace{15pt}\{"Query3": "Calculate the average cost per order for children's food.", "relevant\_column\_list": ["cost"]\},

\hspace{15pt}\{"Query4": "Calculate the average profit per order for children's food.", "relevant\_column\_list": ["profit"]\}

]

**Question**: What is the average concentration of PM2.5 in Sichuan Province in January 2015?\\
**Response**: [

\hspace{15pt}\{"Query1": "Select data from January 2015.", "relevant\_column\_list": ["date<the Gregorian calendar>"]\}, 
    
\hspace{15pt}\{"Query2": "Further filter the data of Sichuan Province from the results of Query1.", "relevant\_column\_list": ["province"]\},
    
\hspace{15pt}\{"Query3": "Calculate the average concentration of PM2.5.", "relevant\_column\_list": ["PM2.5"]\}
]\\
\#\#\# Let's begin!\\
**Table Schema**\\
\{table\_schema\}\\

Response the user's question `\{query\}` strictly follow the above guidelines.\\
**Question**: \{query\}\\
**Response**: \\
\end{tcolorbox}

\subsection{Answer Summary Prompt}

\begin{tcolorbox}[sidebyside, sidebyside align=top seam, width=\linewidth, colback=gray!20, colframe=white, colbacktitle=white, coltitle=white, breakable, arc=0mm, left=0mm, right=0mm]

Based on the following thought process records, generate the Final Answer of the user query "\{query\}" to the table.\\
\#\#\# Rules\\
1. Thoroughly analyze the connection between the query and the thought process, and extract the correct Final Answer.\\
2. Determine the data type of Final Answer based on the understanding of user question. The data type of Final Answer must be one of the following: 

 \hspace{15pt}- Boolean: Valid answers include "True" or "False"(must be string).
 
 \hspace{15pt}- Category: A catgory value (e.g., "Bryin", "try your best!").
 
 \hspace{15pt}- Number: A numerical value, which may represent a computed statistic (e.g., average, maximum).
 
 \hspace{15pt}- List: A list containing number or categories. The expected format example is: ['real estate', 'investments', 'pharmaceuticals', 'software'].\\
3. Output the Final Answer directly without any prefix words or explanations. Your Final Answer's data type must be a number, a category, or a list. Answer with a complete sentences in Final Answer is strictly prohibited.\\

\#\#\# Attention! The data type is just for reference to help you provide the correct format of the Final Answer. The Final Answer content should be derived from the information in the thought process records. Here are your thought process records:\\
\{thought\_process\}\\

=========\\

\#\#\# User Query\\

Query: \{query\}\\

Final Answer: \\
\end{tcolorbox}

\subsection{Iterative Thinking Prompt}

\begin{tcolorbox}[sidebyside, sidebyside align=top seam, width=\linewidth, colback=gray!20, colframe=white, colbacktitle=white, coltitle=white, breakable, arc=0mm, left=0mm, right=0mm]

As an intelligent assistant for table analysis, your primary task is to analyze the table schema and assist in answering questions based on the data. To perform this, follow these guidelines:\\
1.You cannot view the table directly. However, you are provided with schema details and some sample cell values.\\
2.Use these schema details to frame relevant Python queries that progressively solve the user's question.\\
3.Strictly adhere to the structured format below to document your thought process, actions, observations, and responses.\\

**Provided Information**:\\
Schema Retrieval Results:\\
\{table\_schema\}\\

**Thinking Format**:\\
 - Query: Input question that need to be answered. \\
 - Thought: You should always think about what to do and clearly state that.\\
 - Action:  Generate concrete Python-based ideas based on table schema retrieval results to get the observation or answer.\\
 - Observation: Provide observations or results from the action. If unavailable, note the missing information or ambiguities.\\
(Repeat the Thought/Action/Observation steps as needed)\\
 - Thought: After sufficient observations, decide if the original input question can be answered. If so, articulate the response based on the findings.\\
 - Response: Present a concise and accurate answer to the original input question.\\

**Task**:\\
Given the table schema retrieval results above, analyze the input question and generate the thought or response in the structured format.\\

**Input Question**:\\
\{query\}\\

**Thinking Process Records**:\\
\{history\_thinking\}\\

(Remember! Make sure your brief output always adheres to one of the following two formats:\\
A. If the answer to the question can be obtained or inferred from  thinking process records, indicating you have completed the task, please output:\\
**Thought**: 'I have completed the task'\\
**Response**: \\

B. Otherwise, please further rewrite and generate an **improved and clearer query** of the user's target question `\{query\}` based on previous thinking without explanation, and point out potential considerations and error prone points that neeed to be noted, making it easier for LLMs to uderstand and analyse, please output:\\

**Query**: \\
)

\end{tcolorbox}

\subsection{Z-ICL Prompt}

\begin{tcolorbox}[sidebyside, sidebyside align=top seam, width=\linewidth, colback=gray!20, colframe=white, colbacktitle=white, coltitle=white, breakable, arc=0mm, left=0mm, right=0mm]

You are an assistant tasked to response the question asked of a given Table in markdown format. Before providing your response, you need to fully understand and utilize the information contained in the Table. You must response in a single JSON with your answer to the question and your explanation:\\
* "answer": answer using information from the provided Table only.\\
* "explanation": A short explanation on why you gave that answer.\\

\#\#\# Answer Requirements:\\
1. Determine the data type of answer based on the understanding of user question. The data type of answer must be one of the following: \\
 - Boolean: Valid answers include "True" or "False"(must be string).\\
 - Category: A catgory value.\\
 - Number: A numerical value, which may represent a computed statistic (e.g., average, maximum).\\
 - List: A list containing number or categories. The expected format example is: ['real estate', 'investments', 'pharmaceuticals', 'software'].\\
2. Output the response directly without any prefix words or explanations.\\
3. The answer value in response must be derived from the values extracted from the provided data, and any unnecessary rewriting, expansion or format conversion is not allowed. \\

\#\#\# Response Format:\\
Question: What is the name of the richest passenger?\\
Table:

\hspace{15pt}"""

\hspace{15pt}|~passenger~|~wealth(\$)~|

\hspace{15pt}|~------------~|~------------~|

\hspace{15pt}|~value1~|~value2~|

\hspace{15pt}"""\\
Response: \{

    \hspace{15pt}"answer": "value1", 
    
    \hspace{15pt}"explanation": ""
    
\hspace{15pt}\}\\
\\
Now let's start!\\
\\
Question: \{question\}\\
Table: \\
"""\\
\{table\_data\}\\
"""\\
Response: 

\end{tcolorbox}

\subsection{Code-based Prompt}

\begin{tcolorbox}[sidebyside, sidebyside align=top seam, width=\linewidth, colback=gray!20, colframe=white, colbacktitle=white, coltitle=white, breakable, arc=0mm, left=0mm, right=0mm]

You are a professional programming assistant designed to utilize the Python package `pandas` to analyze the table and Response efficient and robust Python code for answering user's Question. The code will read the file from the given `Table\_path` and perform data extraction. 

You should act in accordance with the following requirements:\\
1. Generate chain-of-thought execution ideas based on the understanding of the table content and the user's Question. Describe in detail the algorithm steps as much as possible, including Question analysis, table data format parsing method and code logic description. \\
2. Then write Python codes according to your approach to solve the question. The codes need to be concise and easy to understand, and if necessary, add comments for clarification.\\
3. Note that your analysis must be based entirely on the Table data, with special attention to the content and format of the table cells.\\

You should deliberately go through the user's Question, Table\_path and Table and strictly follow the guidelines to appropriately answer the user's Question. You can only output a standardized JSON object, including "code\_thought" and "code", and you are prohibited from outputting any other unnecessary thought processes. Ensure that your Response can be read by json.loads().\\

\#\#\# Guidelines:\\
**Thought generation**: With the goal of addressing the user's Question, refer to table\_data to generate step-by-step code writing ideas.\\

**File Reading**: Depending on the table file format and size, efficiently read data from the given `Table\_path` (supporting formats such as CSV, Excel) and load it into a Pandas DataFrame. For larger datasets, choose an appropriate method to ensure performance.\\

**String Matching**:\\
- When performing string matching, it is best to use the `.contains()` method instead of a completely strict equal match (`==`). When using `.contains()` function, set the `regex=False`. Usage Example: filtered\_df = df[df['Publication'].str.contains('Harpercollins Publishers (India)', case=False, na=False, regex=False)]\\

**Sorting and Ranking**:\\
- If the query involves rankings, top/bottom N, max/min, higher/lower than, etc, please sort the data using `sort\_values()`. If one or more columns of data to be sorted may have the same value, they should be sorted twice in index order. Usage Example: `df.sort\_values(by='value', ascending=False, kind='mesort')`. \\
- Ensure that the DataFrame is sorted by index even if the values are the same. \\
- When sorting string type numbers, first convert the data type of the numbers from string to float. Usage Example: `sorted\_unique\_ids = sorted([float(u) for u in unique\_supplier\_ids if not pd.isna(u)])\textbackslash n earliest\_5\_suppliers = sorted\_unique\_ids[:5]`\\
- Use unique operation with caution when sorting and ranking.\\

**Special reminders**:\\
- The generated code should be robust, including error handling and file format compatibility. It should strictly match the column names mentioned in the user's Question, avoiding irrelevant or mismatched columns.\\
- Unless otherwise specified, please ignore null or empty values.\\
- Pay attention to the wording of the question to determine if uniqueness is required or if repeated values are allowed. Unless otherwise specified, the unique operation (`.unique()`) is not necessary when sorting or finding the maximum/top/highest/most/lowest/smallest/last (etc) N values in most cases.\\
- Pay attention to **the format of example values** before you manipulate the data in a certain column. Deeply think about how to correctly parse and extract ill-formed data, Not JUST anomaly capture. \\
- For Boolean problems, it is not necessary to output all elements, only obtain True or False answers, or obtain the first few elements to avoid too much unnecessary output.\\
- The results of mathematical operations must be specific number values, and Scientific notation cannot be used.\\

\#\#\# Code Instruction:\\
Your code must be like:\\
"import pandas as pd\textbackslash n def parse\_labels(s):\textbackslash n    if s == '[]':\textbackslash n        return []\textbackslash n    return [label.strip() for label in s.strip('[]').split(',')]\textbackslash n df = pd.read\_csv('all.csv')\textbackslash n \# Explode the labels into individual rows\textbackslash n labels = df['labels\_en'].apply(parse\_labels).explode()\textbackslash n \# Count occurrences of each label\textbackslash n label\_counts = labels.value\_counts()\textbackslash n \# Find the label with the highest number\textbackslash n most\_common\_label = label\_counts.idxmax()\textbackslash n print('the label with the highest number of products', most\_common\_label)"\\

- Ensure the final answer is the last line in python code.\\
- Note that "Answer" is just the placeholder in the code. You should replce it with a entity name or a specific description derived from the user's input, as short as possible. Note that when single quotes are included in the answer description, please use double slashes: `print('Alice's score')`\\

\#\#\# Response Format:\\
**User's Question**: Which label has the highest number of products?\\
Response: \\
\{

\hspace{15pt}"code\_thought":" To find the single label with the highest number of associated products, we'll: 1. Parse the <labels\_en> column to extract individual labels; 2. Handle empty lists and string formatting issues; 3. Count occurrences of each label; 4. Identify the label with the highest count.",

\hspace{15pt}"code": "import pandas as pd\textbackslash n def parse\_labels(s):\textbackslash n    if s == '[]':\textbackslash n        return []\textbackslash n    return [label.strip() for label in s.strip('[]').split(',')]\textbackslash n df = pd.read\_csv('all.csv')\textbackslash n \# Explode the labels into individual rows\textbackslash n labels = df['labels\_en'].apply(parse\_labels).explode()\textbackslash n \# Count occurrences of each label\textbackslash n label\_counts = labels.value\_counts()\textbackslash n \# Find the label with the highest number\textbackslash n most\_common\_label = label\_counts.idxmax()\textbackslash n print('the label with the highest number of products', most\_common\_label)"

\}\\

\#\#\# Let's begin!\\
Now please deliberately go through the following user's Question, Table\_path and Table word by word and strictly follow the above guidelines to appropriately answer the question. You can only output a standardized JSON object, including "code\_thought" and "code", and you are prohibited from responsing without any prefix words or explanations. Ensure that your Response can be read by json.loads().\\

**User's Question**: \{question\}\\
**Table File Path**: \{table\_path\}\\
**Table**:\\
"""\\
\{table\_data\}\\
"""\\
Response: \\
\end{tcolorbox}

\end{document}